\documentclass[pdflatex,sn-mathphys-num]{sn-jnl}

\usepackage{graphicx}%
\usepackage{multirow}%
\usepackage{amsmath,amssymb,amsfonts}%
\usepackage{amsthm}%
\usepackage{amsfonts}
\usepackage{mathrsfs}%
\usepackage[title]{appendix}%
\usepackage{xcolor}%
\usepackage{textcomp}%
\usepackage{manyfoot}%
\usepackage{booktabs}%
\usepackage{algorithm}%
\usepackage{algorithmicx}%
\usepackage{algpseudocode}%
\usepackage{listings}%

\theoremstyle{thmstyleone}%
\theoremstyle{thmstyletwo}%

\theoremstyle{thmstylethree}%

\begin{document}

\title[Article Title]{Multimodal Deep Learning for Low-Resource Settings: A Vector Embedding Alignment Approach for Healthcare Applications}

\author*[1,2]{\fnm{David} \sur{Restrepo}}\email{davidres@mit.edu}\equalcont{These authors contributed equally to this work.}

\author[3]{\fnm{Chenwei} \sur{Wu}}\email{chenweiw@umich.edu}
\equalcont{These authors contributed equally to this work.}

\author[4, 5]{\fnm{Sebastián Andrés} \sur{Cajas}}\email{sebasmos@mit.edu}

\author[1, 6]{\fnm{Luis Filipe} \sur{Nakayama}}\email{luisnaka@mit.edu}

\author[1, 7, 8]{\fnm{Leo Anthony} \sur{Celi}}\email{lceli@mit.edu}

\author[2]{\fnm{Diego M} \sur{López}}\email{dmlopez@unicauca.edu.co}

\affil*[1]{\orgdiv{Laboratory for Computational Physiology}, \orgname{Massachusetts Institute of Technology}, \orgaddress{\city{Cambridge}, \state{Massachusetts}, \country{United States of America}}}

\affil[2]{\orgdiv{Departamento de Telemática}, \orgname{Universidad del Cauca}, \orgaddress{\city{Popayán}, \state{Cauca}, \country{Colombia}}}

\affil[3]{\orgdiv{Department of Electrical Engineering and Computer Science}, \orgname{University of Michigan}, \orgaddress{ \city{Ann Arbor}, \state{Michigan}, \country{United States of America}}}

\affil[4]{\orgdiv{John A. Paulson School of Engineering and Applied Sciences}, \orgname{Harvard University}, \orgaddress{ \city{Boston}, \state{Massachusetts}, \country{United States of America}}}

\affil[5]{\orgdiv{School of Computer Science}, \orgname{University College Dublin}, \orgaddress{ \city{Belfield}, \state{Dublin}, \country{Ireland}}}

\affil[6]{\orgdiv{Department of Ophthalmology}, \orgname{São Paulo Federal University}, \orgaddress{ \city{São Paulo}, \state{São Paulo}, \country{Brazil}}}

\affil[7]{\orgdiv{Department of Biostatistics}, \orgname{Harvard TH Chan School of Public Health}, \orgaddress{ \city{Boston}, \state{Massachusetts}, \country{United States of America}}}

\affil[8]{\orgdiv{Department of Medicine}, \orgname{Beth Israel Deaconess Medical Center}, \orgaddress{ \city{Boston}, \state{Massachusetts}, \country{United States of America}}}


\abstract{
\textbf{Objective:} Large-scale multi-modal deep learning models and datasets have revolutionized various domains such as healthcare, underscoring the critical role of computational power. However, in resource-constrained regions like Low and Middle-Income Countries (LMICs), GPU and data access is limited, leaving many dependent solely on CPUs. To address this, we advocate leveraging vector embeddings for flexible and efficient computational methodologies, aiming to democratize multimodal deep learning across diverse contexts.

\textbf{Background and Significance:} Our paper investigates the computational efficiency and effectiveness of leveraging vector embeddings, extracted from single-modal foundation models and multi-modal Vision-Language Models (VLM), for multimodal deep learning in low-resource environments, particularly in healthcare applications. Additionally, we propose an easy but effective inference-time method to enhance performance by further aligning image-text embeddings.

\textbf{Materials and Methods:} By comparing these approaches with traditional multimodal deep learning methods, we assess their impact on computational efficiency and model performance using accuracy, F1-score, inference time, training time, and memory usage across 3 medical modalities such as BRSET (ophthalmology), HAM10000 (dermatology), and SatelliteBench (public health).

\textbf{Results:} Our findings indicate that embeddings reduce computational demands without compromising the model’s performance, and show that our embedding alignment method improves the performance of the models in medical tasks.

\textbf{Discussion:} This research contributes to sustainable AI practices by optimizing computational resources in resource-constrained environments. It highlights the potential of embedding-based approaches for efficient multimodal learning.

\textbf{Conclusion:}  Vector embeddings democratize multimodal deep learning in LMICs, especially in healthcare. Our study showcases their effectiveness, enhancing AI adaptability in varied use cases.}




\maketitle

\section{Introduction}\label{sec1}

In the era of data-driven decision-making in healthcare, deep learning has emerged as a pivotal methodology for extracting meaningful information from the vast amounts of data from different modalities such as clinical notes, vital signs, lab values, and medical images, among others. The increase of multimodal data, which integrates disparate data formats such as text, image, or audio, requires developing sophisticated computational techniques to process and integrate these heterogeneous data types \cite{3.1, 3.2, c.7}. This integration, known as multimodal data fusion, leverages mainly deep learning techniques \cite{3.3, 3.4} and is critical for building systems that can interpret complex data in a manner akin to human cognition, thereby enhancing decision-making processes in clinical applications \cite{3.5, c.1, c.4, 2.4, 3.7, 2.16, c.6, c.10}: with fundus photos \cite{2.32}, Chest X-rays \cite{3.10}, or even public health applications using remote sensing techniques \cite{3.11, c.8, 3.12, c.5, c.9, 2.78, 3.14, 3.15}.

However, the computational exigencies of such advanced methods pose a formidable barrier, especially in low-resource settings \cite{3.16} environments characterized mainly by limited computational power \cite{3.17, 2.42} and medical data scarcity \cite{3.19, 3.20, c.3, opendata2, c.2, 3.22}. Addressing these constraints requires innovative approaches that optimize computational efficiency without compromising the efficacy of the learning algorithms.

Vector embeddings represent a promising concept in the domain of data efficiency, particularly concerning high-dimensional data like images and text. The embeddings are high-dimensional vectors that encapsulate the essential features of data entities (e.g., words, or images) in a continuous vector space \cite{3.23}. By transforming raw data into a more abstract and computationally manageable form, embeddings facilitate significant reductions in the complexity and dimensionality of data, which is paramount in resource-constrained environments\cite{df-dm}.

The concept of foundation models, representing deep learning architectures primarily based on the transformer framework, has garnered significant attention in recent years \cite{2.19}. These models exhibit remarkable capabilities across diverse domains, such as natural language processing (NLP), exemplified by BERT \cite{2.20}, GPT \cite{3.24}, and LLAMA 2 \cite{2.29}; as well as Computer Vision with Vision Transformer (ViT) \cite{2.22}, or DINO v2 \cite{2.28}; and even multimodal tasks such as Vision Language Models (VLM) with models like CLIP \cite{3.25}, or BLIP 2 \cite{3.26}. Leveraging pre-training on extensive datasets, foundation models offer a versatile starting point for various tasks by providing pre-learned representations that capture a wide array of data characteristics. When applied to multimodal data fusion, embeddings extracted from such models can dramatically lessen the computational load, making it feasible to deploy sophisticated deep-learning models in low-resource settings.

On the other hand, although the use of foundation models such as computer vision models, Large Language Models (LLMs), or VLMs provide us with a robust way of extracting embeddings, we must also take into account that these embeddings may be biased depending on the distribution of data learned during training, mostly from overrepresented groups, perpetuating health biases \cite{2.34}. Liang et al. \cite{3.28} demonstrated that the intrinsic structure of deep learning models generates embedding representations biased to a very small latent space (cone effect). The cone effect generates representations where the image and text embeddings are distant and confined to a very small region of the latent space. Liang et al. also demonstrated how modifying the gap between the embeddings of different modalities improves the fairness and performance of the models \cite{3.28}.

In this paper, we’ll comprehensively examine the computational efficiency gleaned from leveraging vector embeddings extracted from foundation models in multimodal data fusion tasks. We will compare the results with the conventional transfer learning approach using the raw data. Additionally, we’ll demonstrate how the cone effect can be amplified in medical data and propose an embedding alignment method to close the modality gap in medical data. 

This comparison, grounded in a series of methodical experiments across diverse benchmark datasets, aims to elucidate the trade-offs regarding model performance, processing time, memory utilization, and convergence rates. Through this analysis, our research contributes to the broader dialogue on sustainable AI practices, advocating for efficient computational resource utilization in an era marked by the escalating ecological footprint of AI technologies \cite{3.29}.

\section{Methods}\label{sec2}
This study investigates the computational efficiency of using embeddings extracted from foundation models and VLMs versus processing raw data in multimodal deep learning, particularly for healthcare applications. We conducted a series of experiments across three image-text medical datasets to compare the performance of three distinct approaches: 1. Unimodal embedding extraction using a foundation computer vision model, and an LLM for image and text individual embedding extraction; a VLM for image and text embedding extraction; and a transfer learning approach to fine-tune pre-trained transformer-based models using raw data directly. This comparison focuses on key metrics such as accuracy, F1 score, inference time, training time, and memory usage, providing insights into the effective use of computational resources in multimodal data fusion. Additionally, we’ll provide a tool to close the embedding gap between modalities generated in medical data.

\subsection{Datasets}

The evaluations encompassed four multimodal datasets across healthcare applications: diabetic retinopathy, skin lesion classification, and dengue prediction using satellite imagery. The datasets include:

\begin{itemize}
    \item BRSET (Brazilian Multilabel Ophthalmological Dataset) \cite{2.69, 2.74}: A multi-labeled ophthalmological Brazilian dataset. In this case we use BRSET focusing only on binary diabetic retinopathy disease classification. The dataset comprises 16,266 retinal photos from 8,524 patients with metadata corresponding to patient demographic and disease information.

    \item HAM10000 \cite{3.33}: The HAM10000 dataset, an acronym for "Human Against Machine with 10,000 training images," encompasses a comprehensive collection of 10,015 dermatoscopic images for the automated diagnosis of pigmented skin lesions. The images have been sourced from diverse populations and were captured using various dermatoscopic imaging techniques. The categories include Actinic keratoses and intraepithelial carcinoma/Bowen's disease (akiec), basal cell carcinoma (bcc), benign keratosis-like lesions (solar lentigines/seborrheic keratoses and lichen-planus like keratoses, bkl), dermatofibroma (df), melanoma (mel), melanocytic nevi (nv), and vascular lesions (angiomas, angiokeratomas, pyogenic granulomas, and hemorrhage, vasc). 
    
    \item Satellite Images for Public Health (SatelliteBench) \cite{satlmics3, 26}: Adapted from 12-band satellite images to RGB images, the dataset comprises 12,636 satellite images from 81 Colombian municipalities. The task in this dataset involves binary dengue outbreak classification: '1' is assigned to instances with Dengue cases surpassing the median (indicating higher risk), while '0' is assigned to those below. The dataset contains more than 156 images per municipality taken between 2015 and 2018.
    
\end{itemize}

The study used an 80-20 split for training and testing to ensure integrity. Text prompts were generated for datasets without text using prompt templates. with the information of each patient and image. The final number of train and test samples per dataset were defined as:
\begin{itemize}
    \item BRSET \cite{2.69, 2.74}: 13012 training samples, and 3254 testing samples.
    \item HAM10000 \cite{3.33}: 8012 training samples, and 2003 testing samples.
    \item SatelliteBench \cite{satlmics3, 26}: 936 training samples, and 312 testing samples.
    
\end{itemize}

\subsection{Models}

\subsubsection{Single-Modal foundation Models as Embeddings Extractor}
As shown in Figure \ref{fig_embbed:fig1}B, this approach leverages pre-trained foundation models to extract embeddings, hypothesizing that these models, having been trained on extensive datasets, can generate rich, informative representations without further fine-tuning. Image embeddings were obtained using Meta's foundation computer vision model DINO V2 \cite{2.28}, and text embeddings were extracted from Meta's Large Language Model (LLM) LLAMA 2 \cite{2.29}. To alleviate computational resources, for LLAMA 2, we used the smaller version of LLAMA 2-7B with 7 Billion parameters. We also used a version quantized to 8 bits to allow faster inference and lower usage of computational resources. It is important to mention that this same methodology can be applied to bigger models like LLAMA 2-70B, or GPT 4, to extract better-quality information. The embeddings derived from these models were then archived into individual CSV files, ready for subsequent model training and evaluation processes.

\subsubsection{Vision Language foundation Models(VLMs) as Embeddings Extractors}
This method based in Figure \ref{fig_embbed:fig1}B, uses VLMs as embedding extractors, assuming that the model learned a joint representation of both modalities, text, and images. In this case, we selected a CLIP model \cite{3.25}, widely used in the community due to its simplicity and good performance. CLIP was selected also due to its ability to extract uni-modal embeddings instead of a joint embedding representation, allowing us more flexibility during the following experiments. The embeddings extracted from the CLIP model were stored in a CSV file for subsequent model training and evaluation.

\subsubsection{Vector Embeddings Multimodal Fusion Learning}
For the modeling tasks, two fusion techniques, an early (Figure \ref{fig_embbed:fig1}C-1) and late (Figure \ref{fig_embbed:fig1}C-2) fusion methods were employed \cite{3.38}:

\begin{itemize}
    \item Early Fusion: The embeddings from both modalities were concatenated at the input level of our classifier. The classifier consists of a feature extraction block composed of a dense layer with ReLu activation, dropout, and batch normalization. Finally, the output of the previous block was passed to a dense layer with the number of neurons as to the number of output classes for the classification. This approach can be seen in Figure \ref{fig_embbed:fig1}C-1.
    \item Late-Joint Fusion: In the Late-Joint Fusion approach, the embeddings of each modality were passed separately through two feature extraction blocks composed of a dense layer with ReLu activation, a dropout, and a batch normalization of each one. These feature representations were merged later and passed through a final classification head. The approach can be seen in Figure \ref{fig_embbed:fig1}C-2.
\end{itemize}

\subsection{Raw Data Multimodal Fusion Learning}

\begin{figure}[h]
\centering
\includegraphics[width=1\textwidth]{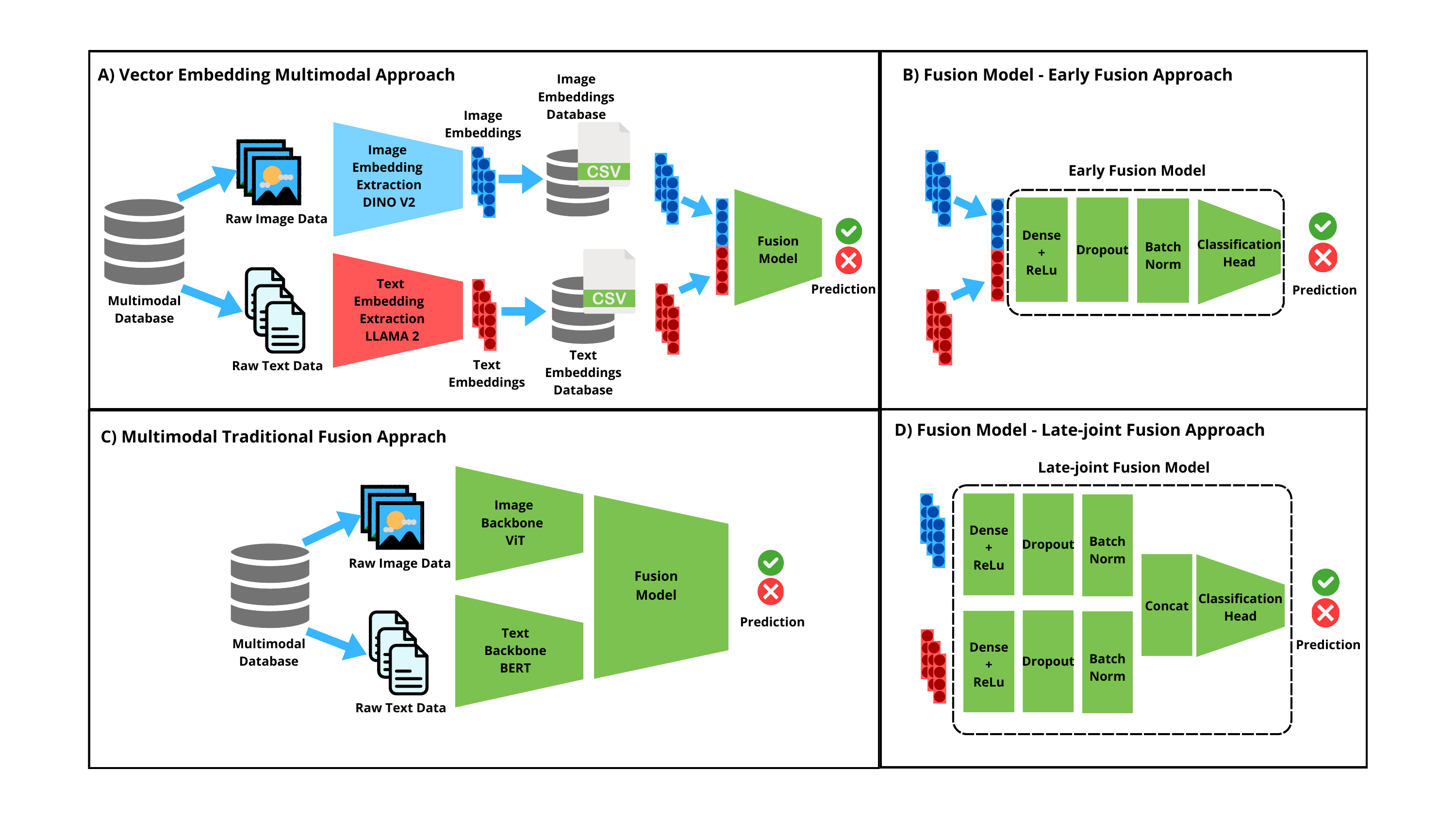}
\caption{Schematic Representation of Multimodal Fusion Approaches. (1A) depicts the traditional multimodal fusion approach using raw data. The approach processes text and images through BERT and ViT models respectively. (1B) shows our embedding multimodal modeling approach, illustrating the extraction of image and text embeddings from foundation models and their subsequent utilization in multimodal learning. (1C) shows the two distinct approaches for data fusion. C-1 represents an early fusion approach, where embeddings are concatenated at the input and passed through a feature extraction block, followed by a classification layer. C-2 presents the late-joint fusion approach, highlighting the separate feature extraction from each modality and their integration at a later stage.}\label{fig_embbed:fig1}
\end{figure}

In this method, represented in Figure \ref{fig_embbed:fig1}A, raw data were directly fed into pre-trained models based on transformers. In this case, a transfer learning approach was used by fine-tuning two transformer-based backbones pre-trained on image and text. Text data were tokenized using a BERT tokenizer and processed through a BERT model architecture [39], while images were inputted into a ViT base architecture [40] pre-trained on ImageNet. The outputs from these models were then integrated using a classification head with the same fusion techniques as the embedding approach in Figure \ref{fig_embbed:fig1}C-1 and Figure \ref{fig_embbed:fig1}C-2, ensuring a consistent comparison between the two methods.

\subsection{Experimental Setup}

\subsubsection{Hardware}
Experiments ran on Oracle's Standard.E4.Flex platform, with 2 CPU cores, 64GB each, no GPU, mimicking low-resource settings. PyTorch executed experiments independently, ensuring consistent model architectures and preventing memory leaks.

\subsubsection{Training and evaluation details}
The AdamW optimizer was employed with default settings from the PyTorch library. We used BCEWithLogitsLoss for binary classification on BRSET, and Satellite Bench and CrossEntropyLoss for multi-class classification on the Ham 10000 dataset. Each loss function was adjusted with class weights inversely proportional to the samples per class to avoid overfitting. A batch size was set to 64 for train and test data loaders, and all the models were trained during 30 epochs. 

For the classifiers, the initial dense layers were set to have 64 neurons in each initial block for late-joint fusion; and 128 neurons in the initial block for early fusion followed by a ReLu activation function. A dropout was defined as 0.0 for the three medical datasets.

Accuracy and F1 score were selected as the performance metrics. The use of F1 as a complement was selected to present the performance of each binary or multi-class classification model while avoiding biased results due to class imbalance. These metrics were reported based on the best test set performance at the end of each epoch during the models' training. In this case, the epoch when the model reached the best performance in the test set, was reported to have a notion of the time to convergence of the model. The models' training and inference times were recorded to compare the model's efficiency in terms of computational resources alongside the memory usage.

To calculate the amount of memory used during training and testing, we iterate over each modality (like 'text', 'images', 'labels') in a batch, determining the memory consumption by multiplying the total number of elements in each tensor by the size of each element and summing these values to get the total memory usage for the batch. For the model, it calculates memory usage by summing the number of elements across all model parameters (weights and biases), and then multiplying by the size of each element, accounting for the data type used (32-bit floats).

\subsubsection{Reducing the Modality Embedding Gap}
The phenomenon known as the "cone effect" in deep neural networks, notably described by Liang et al. \cite{3.28}, highlights how embeddings tend to be concentrated within a narrow region of the high-dimensional space, primarily due to the network's architecture and activation functions. This effect is particularly pronounced in the context of medical data, where the variance in medical images and texts is inherently lower compared to general datasets used for pre-training models. In this section, we provide a formal mathematical demonstration of how the cone effect, induced by random initialization, is amplified in medical datasets. These experiments were carried out for the VLM model CLIP due to its efficiency in performance and efficiency metrics, as well as its ability to extract image and text embeddings independently. To understand how the cone effect in medical data, we need to understand 3 components:

\subsubsection{Increase of Cone Effects in Deeper Models \& Contrastive Models}

Deep learning models are composed of a set of layers, each defined by a non-linear transformation of the input vector 
\( \mathbf{X} = (x_1, x_2, \dots, x_{d_{\text{in}}}) \); \( \mathbf{X} \in \mathbb{R}^{d_{\text{in}}} \).
This transformation involves a linear transformation specified by a weight matrix \( \mathbf{W} \in \mathbb{R}^{d_{\text{out}} \times d_{\text{in}}} \), 
and a bias vector \( \mathbf{b} \in \mathbb{R}^{d_{\text{out}}} \). The resulting linear transformation is given by \ref{eq:linear_transformation}:

\begin{equation}
    \mathbf{Z} = \mathbf{W}\mathbf{X} + \mathbf{b}
    \label{eq:linear_transformation}
\end{equation}

where \( \mathbf{Z} \in \mathbb{R}^{d_{\text{out}}} \). Each linear transformation is followed by a non-linear activation function. One commonly used activation function is the ReLU (Rectified Linear Unit), defined as equation \ref{eq:relu}:

\begin{equation}
    \phi(x) = \max(x, 0)
    \label{eq:relu}
\end{equation}

The ReLU function is applied element-wise to the vector from Equation~\ref{eq:linear_transformation}. The output embedding of a layer in a neural network is thus defined as \ref{eq:output_embedding}:

\begin{equation}
    \text{emb} = \phi(\mathbf{W} \cdot \mathbf{X} + \mathbf{b})
    \label{eq:output_embedding}
\end{equation}

The CLIP model is trained using a contrastive learning approach where they try to minimize the cosine similarity distance between the image embeddings and text embeddings, represented using the equation as:

\begin{equation}
    \mathbf{u} = \phi(\mathbf{W}_{\text{image}}\mathbf{X}_{\text{image}} + \mathbf{b}_{\text{image}})
    \label{eq:image_embedding}
\end{equation}
and
\begin{equation}
    \mathbf{v} = \phi(\mathbf{W}_{\text{text}}\mathbf{X}_{\text{text}} + \mathbf{b}_{\text{text}})
    \label{eq:text_embedding}
\end{equation}
The cosine similarity of the embeddings is defined as equation \ref{cosinesim}:

\begin{equation}
\label{cosinesim}
\cos(u, v) = \frac{\phi(\mathbf{W}_\text{image}\mathbf{X}_\text{image} + \mathbf{b})^\top \phi(\mathbf{W}_\text{text}\mathbf{X}_\text{text} + \mathbf{b})}{\|\phi(\mathbf{W}_\text{image}\mathbf{X}_\text{image} + \mathbf{b})\| \|\phi(\mathbf{W}_\text{text}\mathbf{X}_\text{text} + \mathbf{b})\|}
\end{equation}

In this case, the activation function ensures that all negative components of the vectors are set to zero, thus restricting the vectors to the positive quadrant of the n-dimensional space increasing the similarity in deeper layers as stated in \ref{coneffcos} by Liang et al. \cite{3.28}.

\begin{equation}
\label{coneffcos}
\frac{\phi(\mathbf{W}_\text{image}\mathbf{X}_\text{image} + \mathbf{b})^\top \phi(\mathbf{W}_\text{text}\mathbf{X}_\text{text} + \mathbf{b})}{\|\phi(\mathbf{W}_\text{image}\mathbf{X}_\text{image} + \mathbf{b})\| \|\phi(\mathbf{W}_\text{text}\mathbf{X}_\text{text} + \mathbf{b})\|} \geq  \frac{\mathbf{X}_\text{image}^\top \mathbf{X}_\text{text}}{\|\mathbf{X}_\text{image}\| \|\mathbf{X}_\text{text}\|}
\end{equation}

Additionally, the second part of the CLIP loss is a repulsive structure that further preserves the modality gap \cite{3.41}.

\begin{equation}
-\log \left(\frac{\exp(\text{sim}(x_i, z_i) / \tau)}{\sum_{j=1}^N \exp(\text{sim}(x_i, z_j) / \tau)}\right) = -\text{sim}(x_i, z_i) / \tau + \log \sum_{j=1}^N \exp(\text{sim}(x_i, z_j) / \tau)
\end{equation}

The first term in the equation pulls the positive examples closer, whereas the second term pushes the negative examples away, effectively managing the modality gap.

\subsubsection{Variance Considerations}

\begin{figure}[!htb]
\centering
\includegraphics[width=0.8\textwidth]{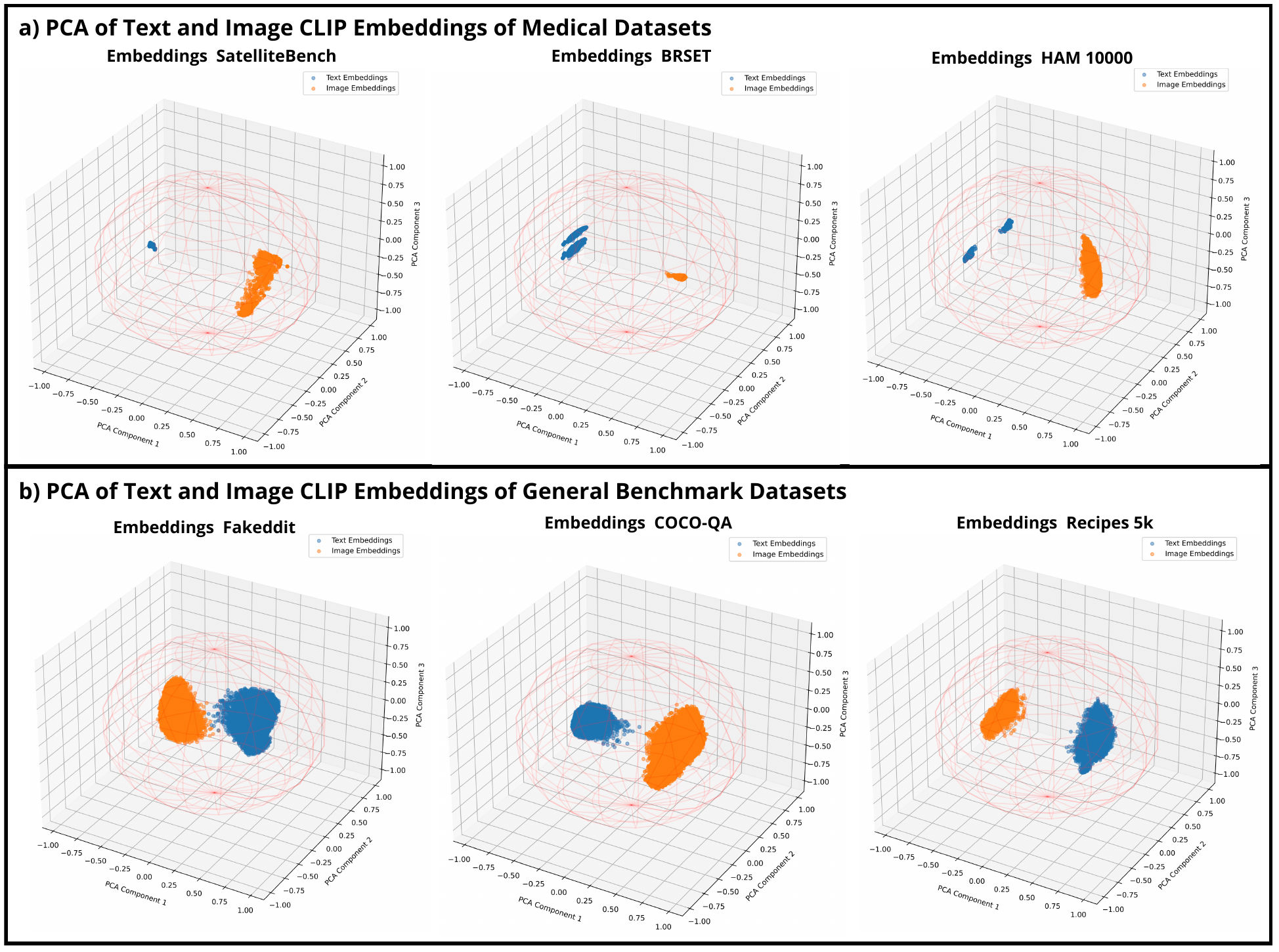}
\caption{Embedding modality gap between image and text embeddings for medical and non-medical datasets. (A) Represents the medical image (orange), and text (blue) embeddings generated using CLIP. (B) Represents the general image (orange), and text (blue) embeddings generated using CLIP from non-medical benchmark datasets. The embedding representations were normalized to fit inside on a unit sphere, and PCA method was used to reduce the dimensionality for visualization.}\label{fig_embbed:fig2}
\end{figure}

Medical data typically exhibit lower variance $\sigma^2_{\text{med}}$ in their embeddings compared to embeddings derived from general natural domain datasets $\sigma^2_{\text{gen}}$ as can be see in \ref{sigma_comp}. This lower variance means that medical data embeddings are more tightly clustered even before any training, making them more susceptible to the cone effect. The reason of this is that, given \( D \), which is the set of all natural domain data and \( M \subseteq D \) represent the medical domain data as a subset. This relationship is expressed as $M \subseteq D$. This effect can be demonstrated empirically when we compare the variance of the embeddings of the 3 medical datasets BRSET \cite{2.69, 2.74} (text embedding = 5.4e-4, image embedding = 7.919e-5), HAM 10000 \cite{3.33} (text embedding = 5.6e-3, image embedding = 6.8e-5), and SatelliteBench \cite{satlmics3, 26} (text embedding = 1.7e-5, image embedding = 3.2e-3) compared with the variance of 3 non-medical benchmarks datasets: COCO-QA \cite{3.42} (text embedding = 3e-3, image embedding = 1.7e-3), Fakeddit \cite{3.43} (text embedding = 1.9e-3, image embedding = 8e-3), and Recipes 5K \cite{3.44} (text embedding = 1.9e-3, image embedding = 8e-3). This difference can also be seen graphically in Figure \ref{fig_embbed:fig2}A and Figure \ref{fig_embbed:fig2}B.

\begin{equation}
\label{sigma_comp}
    \sigma^2_{\text{gen}} \geq \sigma^2_{\text{med}}
\end{equation}

Additionally, as stated by Liang et al. \cite{3.28}, the variance of the hidden state and layer depends directly on the random initialization and the variance of the data, represented as the variance due to the model weights \( \text{Weights}_{\text{variance}} = \operatorname{VAR}[\mathbb{E}[h_{\theta}(\text{embed})]] \) and the variance due to data \( \text{Data}_{\text{variance}} = \mathbb{E}[\operatorname{VAR}[h_{\theta}(\text{embed})]] \). So the total model's variance is represented as:

\begin{equation}
\operatorname{VAR}[h_{\theta}(\text{embed})] = \mathbb{E}[\operatorname{VAR}[h_{\theta}(\text{embed})]] + \operatorname{VAR}[\mathbb{E}[h_{\theta}(\text{embed})]]
\label{eq:total_variance}
\end{equation}

The variance of an intermediate layer in medical contexts can be articulated as:
\begin{equation}
\operatorname{VAR}[h_{\theta}(\text{embed}_{\text{medical}})] = \mathbb{E}[\operatorname{VAR}[h_{\theta}(\text{embed}_{\text{medical}})]] + \operatorname{VAR}[\mathbb{E}[h_{\theta}(\text{embed}_{\text{medical}})]]
\label{eq:medical_variance}
\end{equation}

This contrasts with the variance in more general contexts, where:
\begin{equation}
\operatorname{VAR}[h_{\theta}(\text{embed}_{\text{general}})] = \mathbb{E}[\operatorname{VAR}[h_{\theta}(\text{embed}_{\text{general}})]] + \operatorname{VAR}[\mathbb{E}[h_{\theta}(\text{embed}_{\text{general}})]]
\label{eq:general_variance}
\end{equation}

Finally, since we know that \(\sigma^2_{\text{med}} < \sigma^2_{\text{gen}}\), we can say that:

\begin{equation}
\operatorname{VAR}[h_{\theta}(\text{embed}_{\text{medical}})] < \operatorname{VAR}[h_{\theta}(\text{embed}_{\text{general}})]
\label{eq:variance_comparison}
\end{equation}

\subsubsection{Embedding Alignment \& Shift, Semantic Robustness}
In the preceding discussions, we highlighted the complexities associated with aligning medical image-text data pairs due to the intricate nature of the data and the challenges posed by contrastive learning's deep network requirements and repulsive loss formulation. These factors contribute to a misalignment of extracted embeddings, a problem that not only persists but also intensifies during cross-modal alignment, thereby undermining the robustness of the alignment process. To bridge the modality gap and bolster the semantic robustness of the embedding pairs, we propose the following approach:

\begin{enumerate}
    \item Inject standard normal noise into the embeddings. As each embedding can be viewed as a point on the unit sphere's surface, the addition of Gaussian noise can transform the point into a small sphere. Hence, aligning two embeddings with noise forces the model to acquire the ability to align all the embeddings within the two circles and makes the semantics represented by the circle more robust than the original embedding:
    \begin{align*}
        E_{\text{Text}}' &= E_{\text{Text}} + \theta_t; \\
        E_{\text{Image}}' &= E_{\text{Image}} + \theta_i.
    \end{align*}

    Where $\theta_t \sim \mathcal{N}(0, 1)$ and $\theta_i \sim \mathcal{N}(0, 1)$.

    \item To further refine the cross-modal embedding alignment, we calculate the embedding gap and adjust the embeddings via a shift controlled by the hyperparameter \(\lambda\), followed by renormalization to the unit hypersphere:
    
    \begin{align*}
        \text{Gap} &= \mathbb{E} [\|E_{\text{Text}} - E_{\text{Image}} \| \mid X, Y]; \\
        E_{\text{Text}}' &= E_{\text{Text}} - \frac{\lambda}{2} \times \text{Gap}; \\
        E_{\text{Image}}' &= E_{\text{Image}} - \frac{\lambda}{2} \times \text{Gap}.
    \end{align*}

    \item Moreover, we introduce an additional layer of regularization to the modality alignment process through a hyperparameter-controlled intra-modal alignment loss. This loss function is derived from the outputs of our decoupled late-fusion encoder's branch, aiming to draw paired samples closer and thereby narrow the modality gap. The regularization loss is defined as:
    \begin{equation}
        L_{\text{reg}} = \frac{1}{2N} \sum_{j} \|\| E_{\text{Text}_j}' - E_{\text{Image}_j}' \|\|_2^2
    \end{equation}
\end{enumerate}

\begin{figure}[!htb]
\centering
\includegraphics[width=0.8\textwidth]{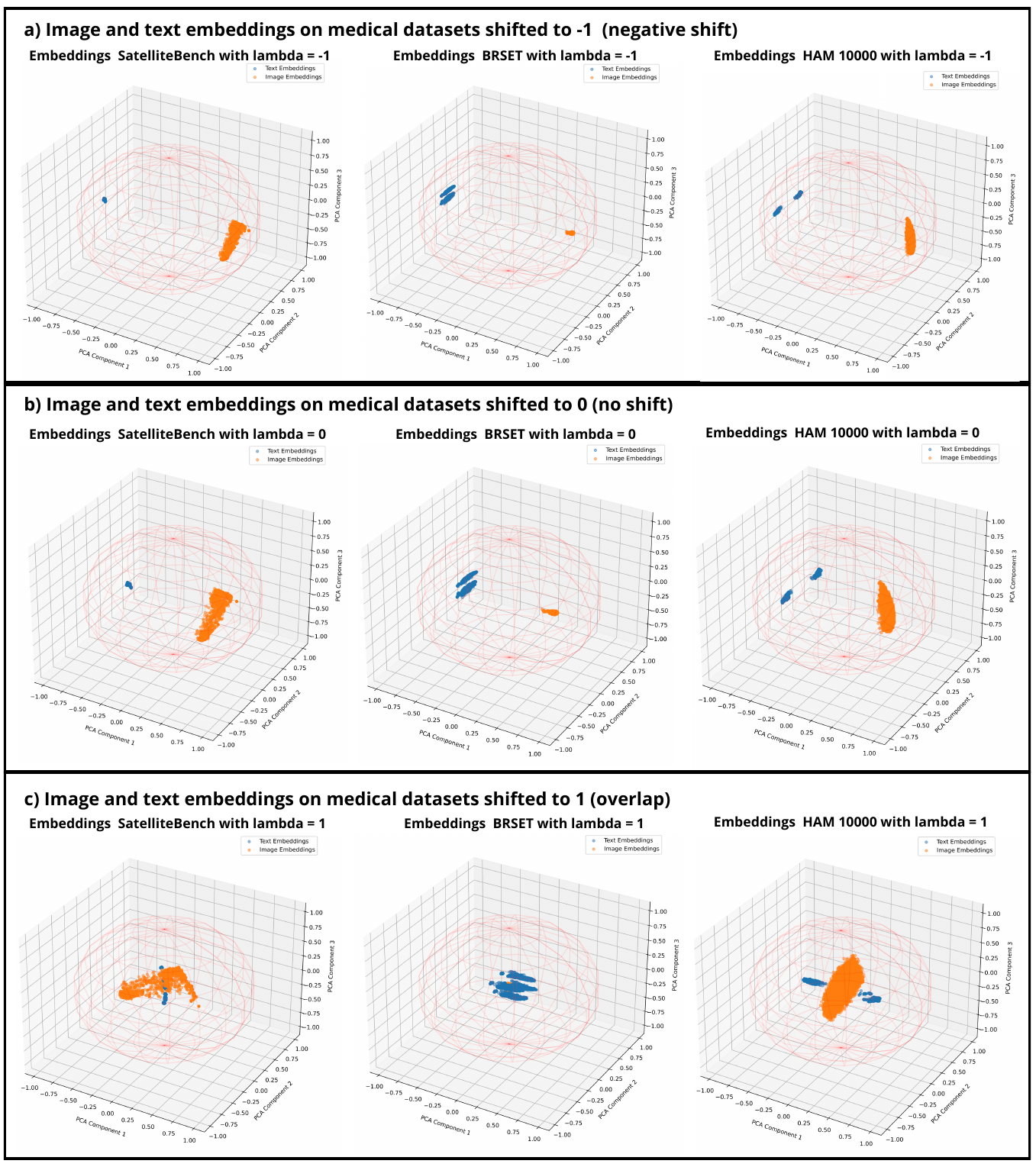}
\caption{Embedding alignment in the medical datasets represented as image embeddings (orange), text embeddings (blue). 3A shows the original embedding representation of each dataset with no shift. 3B Shows the embedding alignment process pooling together both embedding modalities into the same space.}\label{fig_embbed:fig3}
\end{figure}

As result of this process, Figure \ref{fig_embbed:fig3} here visually displays the original embedding representation versus the aligned embeddings.

\section{Results}
Our evaluation extensively demonstrates the performance and efficiency gains achieved by embedding utilization in multimodal deep learning, particularly in resource-constrained environments. Metrics measuring accuracy and computational demands underscore the advantages of embedding-based methods in low-resource scenarios.

\subsection{Performance Metrics}

To calculate the performance metrics of the evaluation of on the three distinct medical datasets, BRSET, HAM 10000, and SatelliteBench, different methodologies were employed to measure their effectiveness in terms of accuracy, F1-score, and convergence epoch. The approaches included embeddings using Dino v2 + Llama 2, CLIP, and direct use of raw data, each tested under early and late-joint fusion methods using our modifications. As can be seen in \ref{tab:performance_metrics}, in the BRSET dataset, the Dino v2 + Llama 2 embedding with early fusion achieved the highest accuracy of 0.987 and an F1-score of 0.944 by the fourth epoch, indicating a rapid convergence and superior model performance. This was closely followed by its late-joint counterpart, which also showed high efficacy with a slightly lower accuracy and F1-score. Similarly, the CLIP approach demonstrated robust performance with both fusion methods, though it peaked slightly later in epochs compared to Dino v2 + Llama 2. The raw data approach lagged behind the embedding methods, indicating the added value of sophisticated feature extraction techniques in handling medical datasets.

\begin{table}[!htb]
\caption{Performance Metrics Across Medical Datasets (Max Epoch 30)}
\label{tab:performance_metrics}
\begin{tabular}{@{}lllccc@{}}
\toprule
\textbf{Dataset} & \textbf{Approach} & \textbf{Method} & \textbf{Accuracy} & \textbf{F1-Score} & \textbf{Epoch} \\ \midrule
\textbf{BRSET} & \textbf{Embedding Dino v2 + Llama 2} & \textbf{Early} & \textbf{0.987} & \textbf{0.944} & \textbf{4} \\
 &  & Late-joint & 0.984 & 0.929 & 14 \\
 & Embedding CLIP & Early & 0.974 & 0.886 & 19 \\
 &  & Late-joint & 0.975 & 0.885 & 14 \\
 & Raw Data & Early & 0.952 & 0.760 & 27 \\
 &  & Late-joint & 0.952 & 0.758 & 8 \\
\midrule
\textbf{HAM 10000} & Embedding Dino v2 + Llama 2 & Early & 0.798 & 0.697 & 28 \\
 &  & Late-joint & 0.815 & 0.715 & 12 \\
 & \textbf{Embedding CLIP} & \textbf{Early} & \textbf{0.818} & \textbf{0.715} & \textbf{21} \\
 &  & Late-joint & 0.811 & 0.712 & 5 \\
 & Raw Data & Early & 0.739 & 0.545 & 8 \\
 &  & Late-joint & 0.743 & 0.617 & 14 \\
\midrule
\textbf{SatelliteBench} & \textbf{Embedding Dino v2 + Llama 2} & Early & 0.752 & 0.751 & 13 \\
 &  & \textbf{Late-joint} & \textbf{0.758} & \textbf{0.758} & \textbf{18} \\
 & Embedding CLIP & Early & 0.734 & 0.733 & 30 \\
 &  & Late-joint & 0.728 & 0.725 & 24 \\
 & Raw Data & Early & 0.574 & 0.570 & 11 \\
 &  & Late-joint & 0.571 & 0.565 & 21 \\
\bottomrule
\end{tabular}
\end{table}

\subsection{Efficiency Metrics}

Table \ref{tab:memory_consumption} provides a comprehensive analysis of the memory efficiency of the different embedding and fusion methodologies. The results delineate a contrast in memory utilization between models that utilize embeddings such as Dino v2 + Llama 2 and CLIP versus those that employ raw data directly. Notably, in the BRSET dataset, the raw data approach consumed substantially more memory (approximately 747.94 MB for model size and over 7471.78 MB per epoch for training data) compared to the Dino v2 + Llama 2 and CLIP methods, which required significantly less memory (2.38 MB and 0.50 MB for model sizes respectively). Similar trends are observed in both HAM 10000 and SatelliteBench datasets where raw data approaches consistently show higher memory footprints, indicating the efficiency of embedding techniques in reducing model and data handling requirements.

\begin{table}[!htb]
\centering
\caption{Memory Consumption Across Medical Datasets (Max Epoch 30)}
\label{tab:memory_consumption}
\begin{tabular}{@{}lllccc@{}}
\toprule
\textbf{Dataset} & \textbf{Approach} & \textbf{Fusion} & \textbf{Model Size} & \textbf{Training} & \textbf{Test} \\ 
& & \textbf{Method} & & \textbf{Dataset Size} & \textbf{Dataset Size}\\
& & & & \textbf{per Epoch} & \textbf{per Epoch}\\ 
\midrule
\textbf{BRSET} & Embedding Dino v2 + Llama 2 & Early & 2.38 MB & 241.48 MB & 15.10 MB \\
 &  & Late-joint & 1.19 MB & 241.48 MB & 15.10 MB \\
 & Embedding CLIP & Early & 0.50 MB & 50.88 MB & 3.18 MB \\
 &  & Late-joint & 0.25 MB & 50.88 MB & 3.18 MB \\
 & Raw Data & Early & 747.94 MB & 7471.78 MB & 467.13 MB \\
 &  & Late-joint & 747.57 MB & 7471.78 MB & 467.13 MB \\
\textbf{HAM 10000} & Embedding Dino v2 + Llama 2 & Early & 2.38 MB & 148.87 MB & 9.45 MB \\
 &  & Late-joint & 1.19 MB & 148.87 MB & 9.45 MB \\
 & Embedding CLIP & Early & 0.50 MB & 31.51 MB & 2.00 MB \\
 &  & Late-joint & 0.25 MB & 31.51 MB & 2.00 MB \\
 & Raw Data & Early & 747.95 MB & 4600.85 MB & 292.12 MB \\
 &  & Late-joint & 747.57 MB & 4600.85 MB & 292.12 MB \\
\textbf{SatelliteBench} & Embedding Dino v2 + Llama 2 & Early & 2.38 MB & 17.37 MB & 1.93 MB \\
 &  & Late-joint & 1.19 MB & 17.37 MB & 1.93 MB \\
 & Embedding CLIP & Early & 0.50 MB & 3.66 MB & 0.41 MB \\
 &  & Late-joint & 0.25 MB & 3.66 MB & 0.41 MB \\
 & Raw Data & Early & 747.94 MB & 537.48 MB & 59.72 MB \\
 &  & Late-joint & 747.57 MB & 537.48 MB & 59.72 MB \\
\bottomrule
\end{tabular}
\end{table}

Aditional to the memory and performance, the table \ref{tab:training_inference_times} highlights the intference and training time. In the table \ref{tab:training_inference_times} we can see how for BRSET the traditional raw data processing required significantly more time both for training (over 538 seconds) and inference (around 134 seconds) per epoch compared to the more advanced embedding techniques using Dino v2 + Llama 2 and CLIP, which drastically reduced these times (ranging from 0.95 to 1.85 seconds for training and 0.28 to 0.72 seconds for inference). Similar trends are observed in the HAM 10000 and SatelliteBench datasets, where raw data approaches consistently consume more computational resources. In particular, the SatelliteBench dataset shows the most substantial efficiency in embedding methods, especially with CLIP, achieving training times as low as 0.16 seconds and inference times around 0.09 seconds per epoch. These results underscore the effectiveness of embedding-based approaches in reducing computational load, thus enhancing the feasibility of deploying these models in real-world applications where quick processing times are crucial.

\begin{table}[!htb]
\centering
\caption{Training and Inference Times Across Medical Datasets (Max Epoch 50)}
\label{tab:training_inference_times}
\begin{tabular}{@{}lllcc@{}}
\toprule
\textbf{Dataset} & \textbf{Approach} & \textbf{Fusion} & \textbf{Average} & \textbf{Average} \\
& & \textbf{Method} & \textbf{Training Time} & \textbf{Inference Time} \\
& & & \textbf{Per Epoch [s]} & \textbf{Per Epoch [s]} \\
\midrule
\textbf{BRSET} & Embedding Dino v2 + Llama 2 & Early & 1.54 & 0.40 \\
 &  & Late-joint & 1.85 & 0.72 \\
 & Embedding CLIP & Early & 0.95 & 0.28 \\
 &  & Late-joint & 1.64 & 0.50 \\
 & Raw Data & Early & 538.38 & 134.14 \\
 &  & Late-joint & 543.11 & 132.89 \\
\midrule
\textbf{HAM 10000} & Embedding Dino v2 + Llama 2 & Early & 1.02 & 0.28 \\
 &  & Late-joint & 1.20 & 0.42 \\
 & Embedding CLIP & Early & 0.65 & 0.20 \\
 &  & Late-joint & 1.12 & 0.42 \\
 & Raw Data & Early & 260.08 & 64.66 \\
 &  & Late-joint & 263.79 & 66.08 \\
\midrule
\textbf{SatelliteBench} & Embedding Dino v2 + Llama 2 & Early & 0.25 & 0.11 \\
 &  & Late-joint & 0.28 & 0.13 \\
 & Embedding CLIP & Early & 0.16 & 0.09 \\
 &  & Late-joint & 0.22 & 0.12 \\
 & Raw Data & Early & 28.64 & 9.63 \\
 &  & Late-joint & 30.34 & 10.12 \\
\bottomrule
\end{tabular}
\end{table}

\subsection{Embedding Alignment}

The embedding alignment was tested by adding variations of the $\lambda$ value to the best performing fusion models on each dataset and plotting the changes on F1-score and accuracy. 

\begin{figure}[!htb]
\centering
\includegraphics[width=0.8\textwidth]{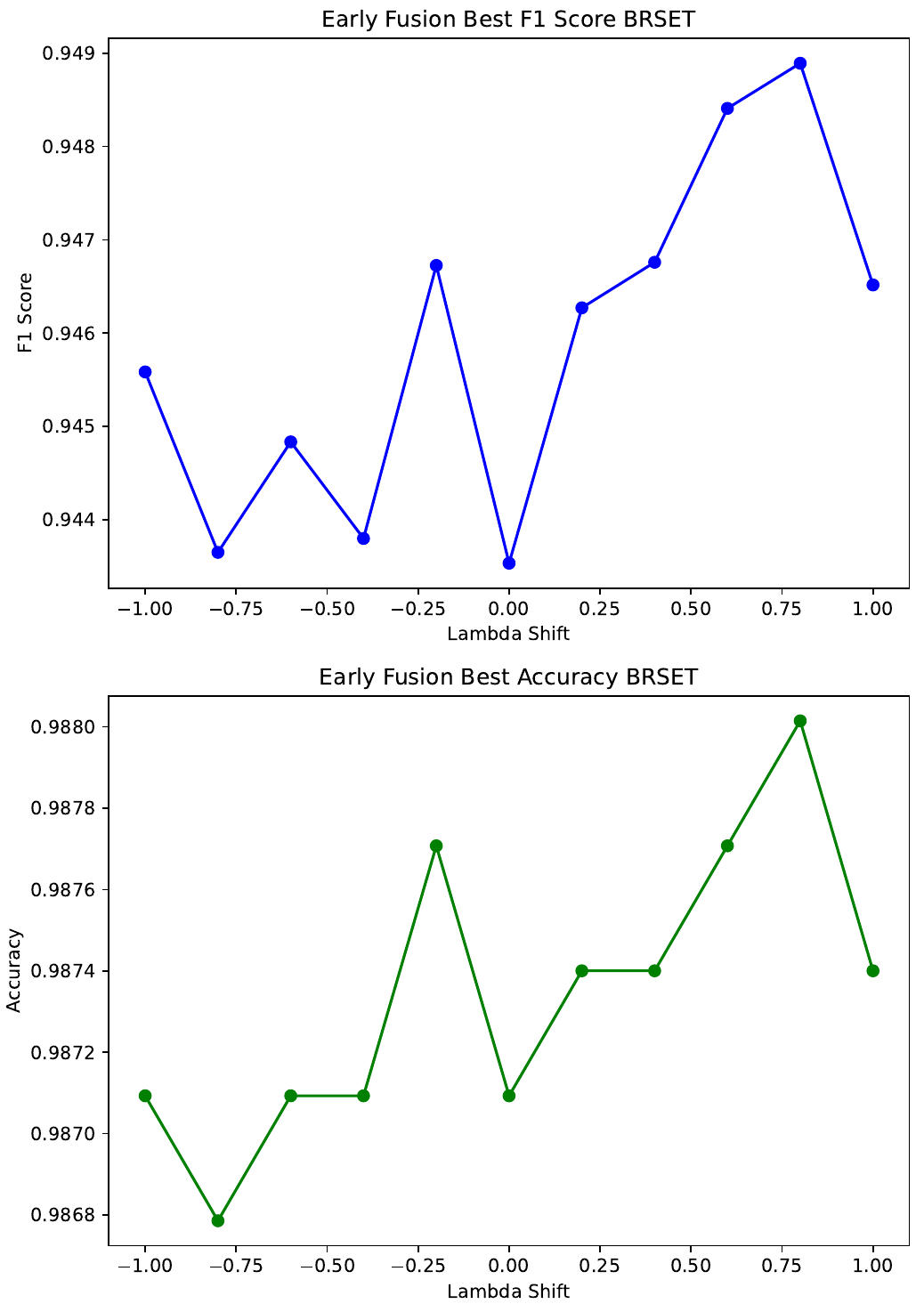}
\caption{Metrics calculated over shifts from negative shift -1, to positive shift 1 for BRSET Dataset.}\label{fig_embbed:fig6}
\end{figure}

The results for BRSET using the early fusion embeddings extracted from Dino v2 and Llama 2 can be seen in figure \ref{fig_embbed:fig6}, where we can see minor improvements in the F1-score from 0.944 to 0.949 using a lambda $\lambda=0.8$, and no substantial change for accuracy.  

\begin{figure}[!htb]
\centering
\includegraphics[width=0.8\textwidth]{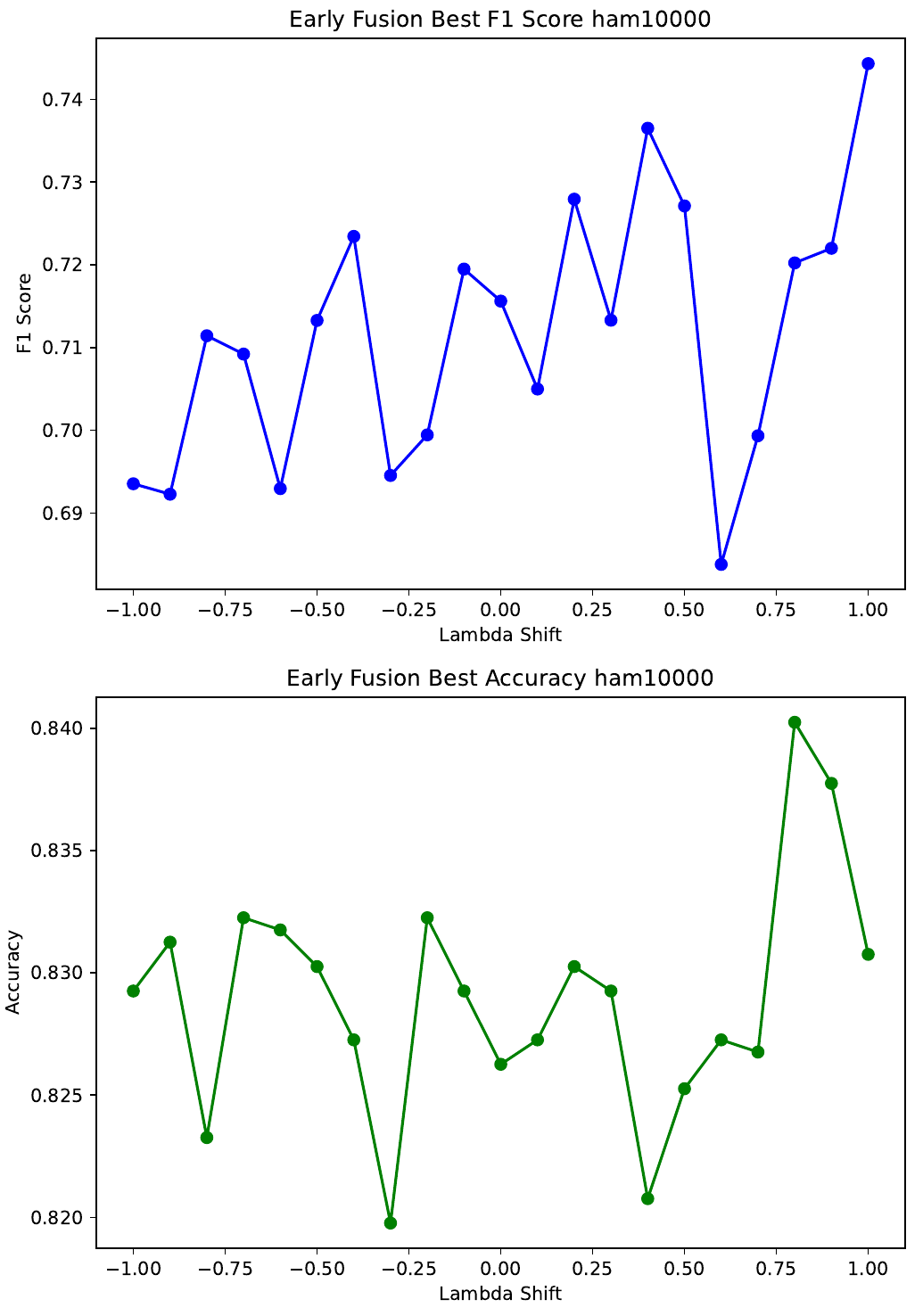}
\caption{Metrics calculated over shifts from negative shift -1, to positive shift 1 for HAM 10000 Dataset.}\label{fig_embbed:fig5}
\end{figure}

\begin{figure}[!htb]
\centering
\includegraphics[width=0.8\textwidth]{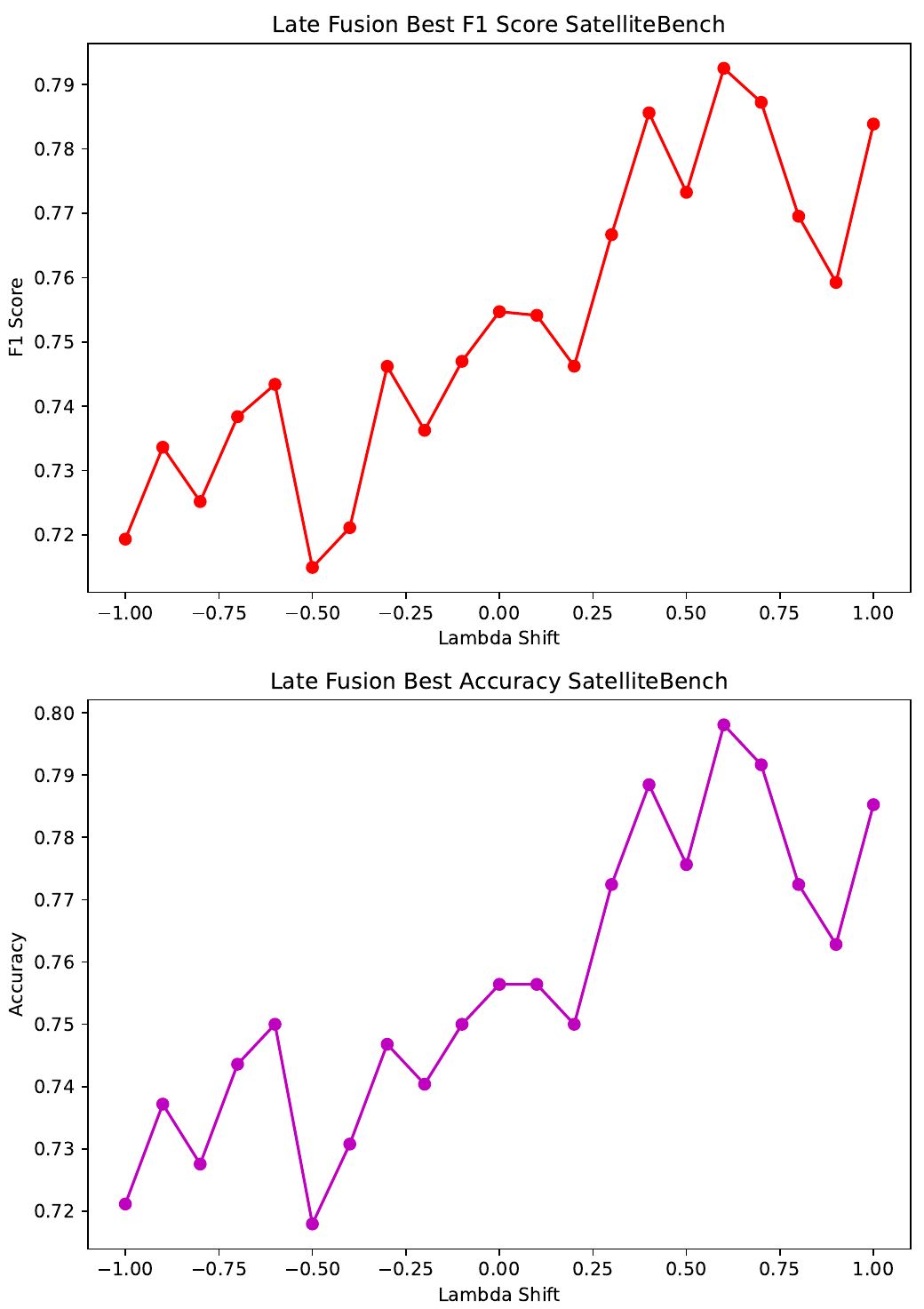}
\caption{Metrics calculated over shifts from negative shift -1, to positive shift 1 for SatelliteBench Dataset.}\label{fig_embbed:fig4}
\end{figure}

The embedding alignment showed an increase in the model performance for SatelliteBench from an F1 score of 0.75 without data shifting, to 0.80 increasing the lambda shifting for the late fusion approach as can be seen in figure \ref{fig_embbed:fig4}. Similar tendency can be seen for the F1 values in Figure \ref{fig_embbed:fig5} for the Ham 10000 dataset where the value of the clip embeddings for early fusion increased from 0.715 with no lambda shift, to 0.745. 

\section{Discussion}
Multimodal vector embeddings present a promising avenue for computationally efficient research, particularly in low-resource settings. Our findings underscore the potential benefits of this approach, notably its simplicity and effectiveness in harnessing the power of pre-trained foundation models without the substantial computational overhead typically associated with fine-tuning or training raw data input models from scratch.

\subsection{Benefits}
The primary advantage of using embeddings lies in their ability to condense complex data into more manageable representations, thereby reducing the computational load and memory requirements. This is particularly beneficial in low-resource environments where computational constraints and specific expertise might limit the deployment of advanced deep-learning models. Our results indicate that embeddings can provide a rich source of pre-encoded information, enabling models to achieve competitive performance levels with significantly less computational demand. This is evident in the reduced training and inference times across all evaluated datasets, highlighting the approach's suitability for real-time applications and environments with limited computational capabilities.

Moreover, the simplicity of the embedding-based approach facilitates ease of implementation and adaptation to various multimodal tasks. By leveraging the generalizability of foundation models like DINO V2 \cite{2.28}  and LLAMA 2 \cite{2.29}, we can extract high-quality embeddings that capture essential features from both images and text, enabling effective multimodal fusion without the need for extensive model customization or hyperparameter tuning.

\subsection{Performance Metrics}
The performance metrics—accuracy and F1-score—demonstrated that the embedding approach generally outperforms the traditional raw data approach in multimodal tasks even using pre-trained models. This superiority can be attributed to the embeddings' ability to condense complex, high-dimensional data into more manageable, semantically rich representations. These compact representations facilitate more efficient learning processes, allowing models to capture the nuances of multimodal data with fewer computational resources.

\subsection{Memory Consumption}
The embedding approach's reduced memory requirements underscore its computational efficiency and practical applicability in low-resource settings. This aspect is crucial for deploying advanced AI models on devices with limited memory capacity, such as mobile devices and embedded systems. Furthermore, the lower memory consumption aligns with sustainable AI practices, reducing the environmental impact associated with data storage and processing.

\subsection{Training and Inference Time Insights}
The significant reduction in training and inference times with the embedding approach directly impacts the practical deployment of deep learning models, especially in real-world scenarios where rapid decision-making is essential. The efficiency gains observed in the study highlight the potential for embeddings to enable advanced deep-learning models on devices with limited computational capabilities, such as mobile phones or embedded systems.

\subsection{Implications}
The discussion extends beyond the immediate findings to consider the broader implications for sustainable AI practices. The embedding approach's ability to deliver competitive performance with reduced computational demands aligns with the growing need for environmentally sustainable AI methodologies. By minimizing the energy and hardware requirements for training and deploying deep learning models, the embedding approach contributes to the development of more eco-friendly AI solutions.

Furthermore, the study's insights into the role of data simplicity and task complexity in model optimization processes underscore the importance of dataset selection and task design in AI research. Understanding how these factors influence model performance and resource efficiency can guide future studies in developing more effective and efficient deep learning algorithms.

\subsection{Limitations}
A notable challenge arises in domain-specific tasks, where the data might significantly deviate from the content typically encountered by foundation models during their training. For instance, in specialized fields such as healthcare, the images and text may encompass highly technical information that is underrepresented in the training corpora of general-purpose models like DINO V2 \cite{2.28} or LLAMA 2 \cite{2.29}. This can result in embeddings lacking crucial domain-specific features, leading to suboptimal performance.

While foundation model embeddings offer rich information for common data, they might miss unique characteristics in specialized datasets. Task-specific models or advanced pre-training techniques like self-supervised learning could address this, albeit with added computational costs, potentially offsetting efficiency gains in general applications.

While the use of embeddings from foundation models offers a compelling strategy for improving computational efficiency in multimodal deep learning, particularly in low-resource settings, it is not a one-size-fits-all solution. The effectiveness of this approach is contingent upon the nature of the task and the characteristics of the data involved. In domain-specific contexts where the data diverge from these norms, alternative strategies, potentially involving task-specific model training or fine-tuning, may be more appropriate. Future research should prioritize developing adaptive, domain-aware embedding strategies and exploring trade-offs between computational efficiency and task-specific performance across various application contexts.

\section{Conclusion}
This paper has presented a comprehensive evaluation of the use of vector embeddings extracted from foundation models for multimodal data fusion in low-resource settings, comparing it against the traditional approach of processing raw data through end-to-end models. The results highlight the potential of using aligned embeddings to significantly reduce the computational burden while retaining, and in some cases enhancing, model performance.

Our findings contribute to the ongoing discourse on sustainable AI practices by offering a viable solution for efficient computational resource utilization. By demonstrating the effectiveness of embeddings in multimodal learning, this work provides a foundation for developing more resource-efficient methodologies in AI, particularly beneficial in resource-limited environments.

However, even when this research shows promising results, further research into task-specific embeddings and advanced pre-training techniques should be carried out. Future work should explore these areas to extend the benefits of efficient multimodal learning across a broader spectrum of domains.

\bmhead{Acknowledgements}

Random Acknowledgements

\section*{Declarations}

The authors declare that they have no conflict of interest.

\end{document}